\documentclass{article}
\usepackage[preprint]{spconf,amsmath,graphicx}
\usepackage{multirow}
\usepackage{multicol}
\usepackage[colorlinks, citecolor=blue]{hyperref}
\usepackage{fancyhdr}
\fancypagestyle{alim}{\fancyhf{}\fancyfoot[c]{\tiny Copyright 2019 IEEE. Published in the IEEE 2019 International Conference on Image Processing (ICIP 2019), scheduled for 22-25 September 2019 in Taipei, Taiwan. Personal use of this material is permitted. However, permission to reprint/republish this material for advertising or promotional purposes or for creating new collective works for resale or redistribution to servers or lists, or to reuse any copyrighted component of this work in other works, must be obtained from the IEEE. Contact: Manager, Copyrights and Permissions / IEEE Service Center / 445 Hoes Lane / P.O. Box 1331 / Piscataway, NJ 08855-1331, USA. Telephone: + Intl. 908-562-3966.}}
\usepackage[numbers,sort&compress]{natbib}

\newlength{\bibitemsep}
\newlength{\bibparskip}
\let\oldthebibliography\thebibliography
\renewcommand\thebibliography[1]{%
	\oldthebibliography{#1}%
	\setlength{\parskip}{\bibitemsep}%
	\setlength{\itemsep}{\bibparskip}%
}

\title{Variational Representation Learning for Vehicle Re-Identification}

\name{\parbox{\linewidth}{\centering
		Saghir Ahmed Saghir Alfasly$^{1,2}$,
		Yongjian Hu$^{1}$, Senior Member, IEEE ,
		Tiancai Liang$^{2}$, 
		Xiaofeng Jin$^{2}$, 
		Qingli Zhao$^{2}$,
		Beibei Liu$^{1}$, Member, IEEE} \thanks{This research has been supported in part by Guangzhou City Science and Technology Foundation under Grant 201710010152, Guangzhou Development District Science and Technology Foundation under Grant 2017GH22, Science and Technology Foundation of Guangdong Province under Grant 2017A050501002 and Grant 2017A030310320, Sino-Singapore Joint Research Institute under Grant 206-A018001 and Grant 206-A017023}} 

\address{ $^{1}$School of Electronic and Information Engineering, South China University of Technology,\\ Guangzhou 510640, P.R.China \\
	$^{2}$GRG Intelligent Security Institute, Guanghzou 510006, P.R.China\\
	$^{1}$\{eeyjhu,eebbliu\}@scut.edu.cn,
	$^{1}$\{eesaghiralfasly\}@mail.scut.edu.cn\\ $^{2}$\{ltcai,jxfeng1,zqli1\}@grgbanking.com}

\begin{document}
\topmargin=0mm 
\maketitle
\thispagestyle{alim}	
\begin{abstract}
Vehicle Re-identification is attracting more and more attention in recent years. One of the most challenging problems is to learn an efficient representation for a vehicle from its multi-viewpoint images. Existing methods tend to derive features of dimensions ranging from thousands to tens of thousands. In this work we proposed a deep learning based framework that can lead to an efficient representation of vehicles. While the dimension of the learned features can be as low as $256$, experiments on different datasets show that the Top-1 and Top-5 retrieval accuracies exceed multiple state-of-the-art methods. The key to our framework is two-fold. Firstly, variational feature learning is employed to generate variational features which are more discriminating. Secondly, long short-term memory (LSTM) is used to learn the relationship among different viewpoints of a vehicle. The LSTM also plays as an encoder to downsize the features.
\end{abstract}
	
	\section{Introduction}
	\label{sec:intro} 
Recently, vehicle image analysis has widely attracted researchers attention due to the increasing demand raised from intelligent public security and public transportation systems. The research mainly focused on vehicle classification  
\cite{compcars,Vehicle_classification_2,4DRDLTDBSV,Vehicle_classification_3}, vehicle detection and tracking
\cite{detection_tracking_1,detection_tracking_2,detection_tracking_3}, vehicle license plate verification \cite{5PROVID} and vehicle retrieval or re-identification \cite{4DRDLTDBSV,5PROVID,VeRi2,2LDNNVRIVSTPP,reid_ret_1,3VPAAMIVRI,6VRISTP,OIF,RAM,reid_ret,reid_ret_2,gan,VRIDHMVI,LCTFSFEVRI,VRIQDDLF}. This paper works on the vehicle re-identification problem. \\ 
The current trend is to turn to deep learning techniques that are believed to be able to automatically derive proper features from various samples. To facilitate the emerging use of deep learning technique, several  well annotated large-scale vehicle datasets \cite{compcars,4DRDLTDBSV,VeRi2,VeRi1} have been published. Two main approaches have been utilized to improve the performance of vehicle re-identification. The first  approach explores the best feature learning strategies for re-identification. Various deep learning models have been used, such as the combination of  generative model with attentive LSTM network in \cite{3VPAAMIVRI}, and the multimodal in \cite{5PROVID}.  Some studies such as \cite{RAM,3VPAAMIVRI,VRIQDDLF} try to improve the output features by increasing their size, while others try to combine several feature vectors from different parallel/stacked sub-models, such as in \cite{5PROVID,2LDNNVRIVSTPP, 3VPAAMIVRI}. There is also work that   combines low level features with high context representation by  mixing hand crafted features with CNN features \cite{5PROVID}.

The second  approach focuses on  the metric learning strategy used for calculating the  similarity or distance between vehicle images.  Representative works include  \cite{5PROVID,3VPAAMIVRI,6VRISTP} that have investigated pairwise distance metric learning ,  \cite{4DRDLTDBSV} that has proposed to learn the set distance metric between positive and negative image sets and  \cite{reid_ret}  that has  utilized triple loss \cite{tripleloss}. The major problem of  existing  methods is that most of them can not be implemented in real-time applications due to the non-negligible large time and space requirement.  What's more, the  discriminating ability of most existing representations are far from perfect. The performance of vehicle re-identification can be absolutely further improved by devising more efficient and accurate representations.
	
\begin{figure*}
	\centering
	\includegraphics[width=18cm]{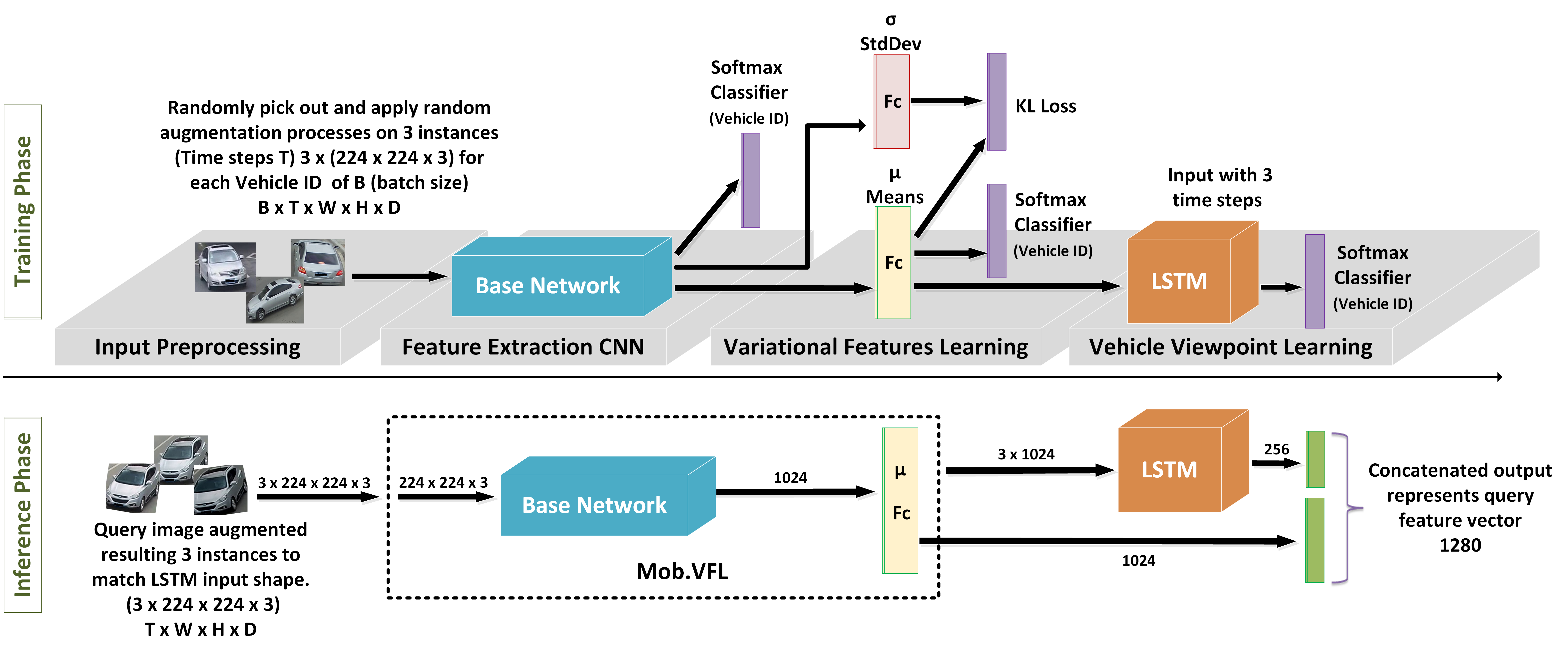}
	\caption{ Training and inference phases of the proposed vehicle re-identificaiton framework. Following the preprocessing, three components are involved, i.e., the CNN feature extractor as the baseline network, the variational feature learning (VFL) network and the LSTM network. The outputs from the second and the last components are jointly used in the inference phase.} 
	\label{fig:model}
\end{figure*}
	
\section{Proposed Model}
\label{sec:model}
Challenges in vehicle re-identification include how to match the query image with the gallery images that are of different viewpoints and, how to distinguish between various vehicle instances of the same color, type or model with the minimum apparence differences. Our framework tried to tackle these problems by combining four stages with three  simple neural networks, which is depicted in Figure \ref{fig:model} and explained in detail as follows.
\subsection{Input Preprocessing and Augmentation}
\label{ssec:input_preprocessing}
The proposed framework has incorporated an LSTM component. Therefore the inputs in training phase are prepared with the dimension \(B\times T \times W \times H \times D \), while inputs dimension in inference phase are prepared with \(T \times W \times H \times D \), where \(B\) is for mini-batch size, \(T\) is for time step, \(W\) and \(H\) are for width and height and \(D\) is for depth of the image sequence, respectively. By default, we have adopted the time step of 3. In the training phase, the time step represents the three different augmented instances randomly picked out from the images of a single vehicle identity . This process helps the training network to learn identical representation for each vehicle regardless of its viewpoint. However, in the inference phase, since there may be only  one query image for the vehicle, we have to generate another two instances by augmentation so as to match the time step of 3,  as it illustrated in the first stage of Figure \ref{fig:model}.  

In order to increase the variability of each vehicle instance images we have applied various image augmentation techniques, including cropping, rotation and brightness adjustment. Based on our experiments, these operations help to increase the performance of vehicle re-identification, while   other operations such as mirroring and color jittering have been excluded due to their negative effect on the training process.

\subsection{CNN Feature Extraction}
\label{ssec:feature_extraction}
A deep convolutional neural network is used in this stage as a baseline network to extract high context features per input instance. Specifically, we have employed the MobileNet v1 \cite{mobilenet} that not only helps to reduce overfitting but also runs faster than regular CNN with much fewer parameters. The output of this stage is a feature vector of size $1024$ for each time step.
\subsection{Variational Feature Learning}
\label{ssec:variational_learning}
Inspired by the variational autoencoder learning \cite{va}, we trained the model with KL (Kullback-Leibler) divergence. Two fully connected layers are used to predict the means \(\mu\) and the standard deviation \(\sigma\) of a Gaussian distribution \(\mathcal{N}(\mu,\sigma)\) . The outputs of this stage ensure the representation to diverse sufficiently. Besides, the outputs $\mu$ is well normalized. The KL divergence between the \(\mu,\sigma\) distribution and the prior is considered as a regularization which helps to overcome the overfitting problem. Along with KL divergence, Softmax classifier on top of the means \(\mu\) layer is used to learn this network as it illustrated in third stage of Figure \ref{fig:model}. 

\subsection{Viewpoint Learning and Encoding}
\label{ssec:viewpoint_encoding}
While LSTM is normally used to learn the temporal relationship between a sequence of samples, we utilize it in this stage to achieve two goals: vehicle viewpoint understanding and feature encoding. To this end, we feed the LSTM with \(T\) different instances, possibly of different viewpoints, of the same vehicle identity. It is expected that similar feature representation for the same vehicle regardless of the viewpoints can be learned. Another role of this layer is that we can further reduce the size of output features by controlling how many units are used in the LSTM.  We have tested with different numbers of units and the results were quite promising that the performance did not drop significantly with the feature size reducing to as low as $64$.   Finally, on LSTM layer is trained using Softmax classifier  on top of it with vehicle ID labels. Note that in the inference phase, the final vehicle feature vector is the  output sequence of LSTM rather than the label vector.

\section{Training and Inference}
\label{sec:training}
The models described in section \ref{sec:model} can be trained in different ways, either jointly or separately, depending on the characteristics of the datasets. In this work, for the VeRi dataset, we empirically found that the best practice is to train the three components separately, i.e., we freeze the previous components when training the current component.  
For the VehicleID dataset, since the third component of LSTM is not required due to the limited number of viewpoints, we have trained the CNN and the VFL components jointly. We fine-tuned pretrained MobileNet \cite{mobilenet} on ImageNet \cite{imagenet} for both datasets.  The first and the last components are trained  with the Softmax classifier  for 70 epochs using Adam optimizer \cite{Adam} , with the initial learning rate set to 0.0001 and multiplied by 0.1 every 30 epochs. The equation (\ref{eq:softmax}) defines the Softmax classifier where \(v\) denotes the output vector, \(t\) denotes the target vector, and \(p_j\) denotes the input to the neuron \(j\). 

\begin{equation}
\label{eq:softmax}
\begin{aligned}
L_{id}(t,v)= -\sum^v_{i=1} t_i log\bigg(\dfrac{e^{p_i}}{\sum^v_j e^{p_j}}\bigg)
\end{aligned}
\end{equation}

For the second component, i.e., the VFL network,  we used two weighted loss functions, as is shown in equation (\ref{eq:all}).

\begin{equation}
\label{eq:all}
\begin{aligned}
L(t,v,\mathcal{N}(\mu,\sigma),\mathcal{N}(0,1))= L_{id}(t,v) + \\\alpha D_{kl}(\mathcal{N}(\mu,\sigma),\mathcal{N}(0,1))
\end{aligned}
\end{equation}

 The second term  \(D_{kl}\) represents the KL loss and is defined by equation (\ref{eq:kl}). The weight of KL loss \(\alpha\) is    empirically set to $(0.1)$ in this work.  \(\mu\) and \(\sigma\) denote the predicted means and standard deviation of the Gaussian distribution while  \(n\) denotes the vector size.  This component is trained for 50 epochs using Adam optimizer , with the initial learning rate set to 0.0001 and multiplied by 0.1 every 20 epochs. 
\begin{equation}
\label{eq:kl}
\begin{aligned}
D_{kl}(\mathcal{N}(\mu,\sigma),\mathcal{N}(0,1))=-\dfrac{1}{2}\sum^n_{i=1}\bigg( 1+ log(\sigma_i)\\ - \mu^2_i - \sigma_i\bigg)
\end{aligned}
\end{equation}
\begin{table}[htb]
	\centering
	\caption{Model Components Performance on VeRi.}
	\label{VeRi_componenets}      
	\begin{tabular}{ccccccc}
		\hline \noalign{\smallskip}
		Component   & Output Length  & Top-1 &  Top-5\\
		\hline\noalign{\smallskip}
		\multirow{5}{1.5cm}{Mob.LSTM} 
		& 64 & 75.20 & 88.25\\
		& 128 & 81.64 & 90.28 \\
		& 256  & 83.37 & 92.37 \\
		& 512  & 83.61 & 90.94\\
		& 1024  & 83.19 & 91.06\\
		\hline\noalign{\smallskip}
		\multirow{2}{1.5cm}{Mob.VFL} 
		& 256  & 85.28 & 92.49\\
		& 1024 & 85.63 & 92.66\\
		\hline
	\end{tabular}
\end{table}
\begin{table}[htb]
	\begin{center}
		\caption{Vehicle Re-Identification Comparison on VeRi}      
		\begin{tabular}{ccccc}
			\hline\noalign{\smallskip}
			Model  & mAP \% & Top-1 &  Top-5\\
			\hline\noalign{\smallskip}
			Siamese-Visual \cite{2LDNNVRIVSTPP} & 29.48  & 41.12 & 60.31 \\
			FACT \cite{VeRi1} & 18.49  & 50.95 & 73.48 \\
			XVGAN \cite{gan} & 24.65  & 60.20 & 77.03 \\
			OIF \cite{OIF} & 48.00  & 65.92 & 87.66 \\
			Siamese-CNN \cite{2LDNNVRIVSTPP} & 54.21  & 79.32 & 88.92 \\
			VAMI \cite{3VPAAMIVRI} & 50.13  & 77.03 & 90.82\\
			Path-LSTM \cite{2LDNNVRIVSTPP} & 54.49  & 82.89 & 89.81 \\
			VR-PROUD \cite{reid_ret_2} & 40.5  & 83.19 & 91.12 \\ 
			D-DLF \cite{VRIQDDLF} & 53.26  & 84.92 & 93.03 \\
			Mob.VFL-LSTM(Ours) &  \textbf{58.08} &  \textbf{87.18} &  \textbf{94.63}\\
			\hline\noalign{\smallskip}
			SCCN-Ft+CLBL-8-Ft \cite{VRIDHMVI} &  25.12  & 60.83 & 78.55\\
			\multirow{2}{3.8cm}{\centering FACT+Plate-SNN+ STR \cite{VeRi2}} & \multirow{2}{1cm}{\centering 27.77}  & \multirow{2}{1cm}{\centering 61.44} & \multirow{2}{1cm}{\centering 78.78} \\\\
			OIF + ST \cite{OIF} & 51.42  & 68.30 & 89.70 \\
			\multirow{2}{3.8cm}{\centering Siamese-CNN+Path-LSTM\cite{2LDNNVRIVSTPP}} & \multirow{2}{1cm}{\centering 58.27}  & \multirow{2}{1cm}{\centering 83.49} & \multirow{2}{1cm}{\centering 90.04} \\\\
			VAMI \cite{3VPAAMIVRI} + STR \cite{VeRi2} & \textbf{61.32}  & 85.92 & 91.84\\
			QD-DLF \cite{VRIQDDLF} & 61.83  & \textbf{88.50} & 94.46 \\ 
			\multirow{2}{3cm}{\centering Mob.VFL-LSTM + Mob.VFL$^{\star}$ (Ours)} & \multirow{2}{1cm}{\centering 59.18}  &  \multirow{2}{1cm}{\centering 88.08} &  \multirow{2}{1cm}{\centering \textbf{94.63}}\\\\
			\hline
		\end{tabular}
		\label{veri}
	\end{center}
\end{table} 
\begin{table*}[htb]
	\begin{center}
		\caption{Comparison of Different Vehicle Re-Identification Methods on VehicleID }
		\label{vehicle_id_comp}
		\begin{tabular}{ccccccccc}
			\noalign{\smallskip}\hline\noalign{\smallskip}
			\multirow{2}{2cm}{ Model}  & \multicolumn{2}{c}{ Test size 800} && \multicolumn{2}{c}{ Test size 1600} && \multicolumn{2}{c}{ Test size 2400} \\
			\cline{2-3}
			\cline{5-6}
			\cline{8-9}
			\noalign{\smallskip}
			&Top-1&Top-5&&Top-1&Top-5&&Top-1&Top-5\\ \hline\noalign{\smallskip}
			VGG-Triplet Loss\cite{4DRDLTDBSV} & 40.40 & 61.70 && 35.40 & 54.60 && 31.90 & 50.30 \\
			VGG-CLL\cite{4DRDLTDBSV} & 43.60 & 64.20 && 37.00 & 57.11 && 32.90 & 53.30 \\
			GoogLeNet \cite{compcars} & 47.88 & 67.18 && 43.40 & 63.86 && 38.27 & 59.39 \\
			FACT \cite{VeRi1} & 49.53 & 68.07 && 44.59 & 64.57&& 39.92 & 60.32 \\
			Mixed Diff-CLL\cite{4DRDLTDBSV}  & 49.00 & 73.50 && 42.80 & 66.80 && 38.20 & 61.60 \\
			XVGAN \cite{gan}  & 52.87 & 80.83 && 49.55 & 71.39 && 44.89 & 66.65  \\
			C2F-Rank \cite{LCTFSFEVRI}  & 61.10 & 81.70 && 56.20 & 76.20 && 51.40 & 72.20  \\
			VAMI \cite{3VPAAMIVRI}  & 63.12 & 83.25 && 52.87 & 75.12 && 47.34 & 70.29  \\
			Mob.VFL$^{\star}$ (Ours) & \textbf{73.37} & \textbf{85.52} && \textbf{69.52} & \textbf{81.00} && \textbf{67.41} & \textbf{78.48} \\    
			
			\hline
		\end{tabular}
	\end{center}
\end{table*}
In the inference phrase, as illustrated in  Figure \ref{fig:model}, we only keep the Means $\mu$ layer for the VFL component. The overall inference pipeline involves the  CNN feature extractor, the Mean $\mu$ layer and the LSTM layer. Note that we concatenate the outputs from both the last and the second components as the final feature vector.
\section{Experimental Results}
\label{sec:evaluation}
We have evaluated our method on two large-scale public datasets  VehicleID \cite{4DRDLTDBSV} and VeRi \cite{VeRi1}. The VeRi dataset contains $37,778$ images of $576$ vehicles for training and $11,579$ images of $200$ vehicles for testing. The pictures of this dataset are taken from various viewpoints with several surveillance cameras for each vehicle.
 
We have first done extensive experiments to investigate the effect of individual components as well as the impact of different feature sizes. As is shown in Table \ref{VeRi_componenets}, for the combination of  base CNN network with LSTM (denoted as Mob.LSTM), the Top-1 and Top-5 accuracies differ with the output sequence lengths. The best results are achieved when the length is 256. It is worth noting that the results have not dropped dramatically when the length is reduced to as low as 64, demonstrating the effectiveness of using LSTM in the model. On the other hand, the outputs of variational feature learning network (denoted as Mob.VFL) are slightly better than that of Mob.LSTM, with the highest accuracies being achieved when the output feature size is 1024.

To further verify the advantage of our work, we have compared our methods with several most recent related works, including Siamese-Visual \cite{2LDNNVRIVSTPP}, FACT \cite{VeRi1}, XVGAN \cite{gan}, OIF \cite{OIF}, Siamese-CNN \cite{2LDNNVRIVSTPP},  VAMI \cite{3VPAAMIVRI}, Path-LSTM  \cite{2LDNNVRIVSTPP}, VR-PROUD \cite{reid_ret_2}, and D-DLF \cite{VRIQDDLF}.  As is shown on the upper part of Table \ref{veri}, our method provides the best Top-1 and Top-5 performance compared with the state-of-the-art methods.
 
It is generally believed that combining multiple networks may possibly boost the performance. For example, in \cite{2LDNNVRIVSTPP} the authors have combined Siamese-CNN with Path-LSTM and observed a boost on the performance. In \cite{VRIQDDLF}, the authors combined $4$ CNN network instances with different directional pooling layers to improve the performance. We therefore have used the default configuration of our method involving all three components as the base network and used the combination of CNN  and the VFL components (jointly trained and denoted as Mob.VFL$^{\star}$ ) as the second network. We then concatenate the outputs of these two networks and compare the results with other similar practices from the literature, as listed in the lower part of Table \ref{veri}. It is shown that the combination of different configurations of our models has achieved the best performance in terms of Top-5.

VehicleID contains $113,346$ images of $13,164$ vehicles for training and $108,221$ images of $13,164$ vehicles for testing. This dataset contains only two viewpoints front and rear for each vehicle, which forces us to train without the LSTM layer but only with the baseline network and VFL jointly. We followed the evaluation protocol proposed in \cite{4DRDLTDBSV} and worked on three testing sets of size $800$, $1600$ and $2400$ respectively. 
Table \ref{vehicle_id_comp} shows the comparison between our method and several latest leading methods in the literature. Among these methods, VAMI of \cite{3VPAAMIVRI}, which is a complex model using attention mechanism with generative adversarial network (GAN), shows the best performance in terms of the Top-1 and Top-5  retrieval accuracy. However, our method (denoted as Mob.VFL$^{\star}$) has achieved even better results than VAMI on all three test sets.  Particularly, in the biggest testing set of 2400, our method has significantly increased the Top-1 accuracy from $47.34$ of VAMI to $67.10$.

\section{Conclusion}
\label{sec:conclusion}
In this work, we have devised an efficient deep-learning-based framework that can derive highly discriminating representations for vehicle images, which helps to improve the performance of vehicle re-identificaton. Our framework is characterized by the combination of three components dedicated to different purposes. The base CNN network extracts the high context features based on which, the following variational feature learning component learns Gaussian distribution of the same vehicle instances. The last component which involves an LSTM layer plays two roles:  extracting intra-class features from different viewpoints and reducing the size of the output feature vectors.

The effectiveness of our framework has been demonstrated by extensive experiments. It is also believed that the idea of using  variational feature learning  with KL divergence can not only boost the performance of vehicle re-identificaton, but also  be extended  to other similar scenarios such as content-based image retrieval and fine-grained classification to improve the quality of object representation. 

\bibliography{variational-representation-learning-for-vehicle-reid-preprint.bbl}

\end{document}